\documentclass[sigconf]{acmart}

\AtBeginDocument{%
  }

\copyrightyear{2023}
\acmYear{2023}
\setcopyright{acmlicensed}\acmConference[WWW '23]{Proceedings of the ACM Web Conference 2023}{May 1--5, 2023}{Austin, TX, USA}
\acmBooktitle{Proceedings of the ACM Web Conference 2023 (WWW '23), May 1--5, 2023, Austin, TX, USA}
\acmPrice{15.00}
\acmDOI{10.1145/3543507.3583498}
\acmISBN{978-1-4503-9416-1/23/04}

\usepackage[utf8]{inputenc} % allow utf-8 input
\usepackage[T1]{fontenc}    % use 8-bit T1 fonts
\usepackage{hyperref}       % hyperlinks
\usepackage{url}            % simple URL typesetting
\usepackage{booktabs}       % professional-quality tables
\usepackage{amsfonts}       % blackboard math symbols
\usepackage{nicefrac}       % compact symbols for 1/2, etc.
\usepackage{microtype}      % microtypography
\usepackage{xcolor}         % colors
\usepackage{comment}
\usepackage{multirow}
\usepackage{amsmath}
\usepackage{graphicx}
\usepackage{pifont}
\usepackage{makecell}
\usepackage{amsthm}
\usepackage{bm}
\usepackage{algorithm}  
\usepackage{algorithmic}
\usepackage{enumitem}
\newcommand{\cmark}{\ding{51}}
\newcommand{\xmark}{\ding{55}}

\usepackage{wrapfig,lipsum,booktabs}
\usepackage{caption}
\usepackage{setspace}
\usepackage[toc,page]{appendix}
\def\eg{\emph{e.g.}}

\def\ie{\emph{i.e.}}

\def\vs{\emph{vs}}
 
\begin{document}

%%
%% The "title" command has an optional parameter,
%% allowing the author to define a "short title" to be used in page headers.
\title{Auto-HeG: Automated Graph Neural Network on Heterophilic Graphs}

%%
%% The "author" command and its associated commands are used to define
%% the authors and their affiliations.
%% Of note is the shared affiliation of the first two authors, and the
%% "authornote" and "authornotemark" commands
%% used to denote shared contribution to the research.

\author{Xin Zheng}
\authornotemark[5]
\email{xin.zheng@monash.edu}
\affiliation{%
  \institution{Monash University}
  \city{Melbourne}
  %\state{VIC}
  \country{Australia}
}
\author{Miao Zhang}
\authornotemark[2]
\email{zhangmiao@hitsz.edu.cn}
\affiliation{%
  \institution{Harbin Institute of Technology (Shenzhen)}
  \city{Shenzhen}
  %\state{Guangdon}
  \country{China}
}
\author{Chunyang Chen}
\authornotemark[5]
\email{chunyang.chen@monash.edu}
\affiliation{%
  \institution{Monash University}
  \city{Melbourne}
%  \state{VIC}
  \country{Australia}
}
\author{Qin Zhang}
\authornotemark[3]
\email{qinzhang@szu.edu.cn}
\affiliation{%
  \institution{Shenzhen University}
  \city{Shenzhen}
  %\state{Guangdon}
  \country{China}
}
\author{Chuan Zhou}
\authornotemark[4]
\email{zhouchuan@amss.ac.cn}
\affiliation{%
  \institution{Chinese Academy of Sciences}
  \city{Beijing}
  \country{China}
}
\author{Shirui Pan}
%\authornotemark[5]
\authornote{Corresponding author.}
\email{s.pan@griffith.edu.au}
\affiliation{%
  \institution{Griffith University}
  \city{Gold Coast}
  %\state{QLD}
  \country{Australia}
}

\begin{comment}
\author{Xin Zheng$^{1}$,~~~Miao Zhang$^{2}$,~~~Chunyang Chen$^{1}$,~~~Qin Zhang$^{3}$,~~~Chuan Zhou$^{4}$,~~~Shirui Pan$^{5}$*}

\makeatletter
\def\authornotetext#1{
 \g@addto@macro\@authornotes{%
 \stepcounter{footnote}\footnotetext{#1}}%
}
\makeatother

\authornotetext{Corresponding author}

\affiliation{%
 \institution{$^1$Monash University \quad $^2$Harbin Institute of Technology (Shenzhen) \quad $^3$Shenzhen University \\ $^4$ Chinese Academy of Sciences \quad $^5$Griffith University}
 \country{}
}

\email{{xin.zheng, chunyang.chen}@monash.edu;zhangmiao@hitsz.edu.cn;}
\email{qinzhang@szu.edu.cn;  
zhouchuan@amss.ac.cn; 
s.pan@griffith.edu.au}
\end{comment}
%%
%% By default, the full list of authors will be used in the page
%% headers. Often, this list is too long, and will overlap
%% other information printed in the page headers. This command allows
%% the author to define a more concise list
%% of authors' names for this purpose.
\renewcommand{\shortauthors}{Zheng et al.}

%%
%% The abstract is a short summary of the work to be presented in the
%% article.
\begin{abstract}
Graph neural architecture search (NAS) has gained popularity in automatically designing powerful graph neural networks (GNNs) with relieving human efforts. 
However, existing graph NAS methods mainly work under the homophily assumption and overlook another important graph property, \ie, heterophily, which exists widely in various real-world applications.
%However, existing graph NAS methods mainly work under the homophily assumption. 
%In contrast, heterophily, which shares an opposite property of graph data to homophily, exists widely in various real-world applications, \eg, online social networks and transactions. 
%Despite its vital role in the web socio-economic system.
To date, automated heterophilic graph learning with NAS is still a research blank to be filled in.  
Due to the complexity and variety of heterophilic graphs, the critical challenge of heterophilic graph NAS mainly lies in developing the heterophily-specific search space and strategy. 
Therefore, in this paper, we propose a novel automated graph neural network on heterophilic graphs, namely {\bf{Auto-HeG}}, to automatically build heterophilic GNN models with expressive learning abilities.
Specifically, Auto-HeG incorporates heterophily into all stages of automatic heterophilic graph learning, including search space design, supernet training, and architecture selection.
Through the diverse message-passing scheme with joint micro-level and macro-level designs, we first build a comprehensive heterophilic GNN search space, enabling Auto-HeG to integrate complex and various heterophily of graphs. 
With a progressive supernet training strategy, we dynamically shrink the initial search space according to layer-wise variation of heterophily, resulting in a compact and efficient supernet.
Taking a heterophily-aware distance criterion as the guidance, we conduct heterophilic architecture selection in the leave-one-out pattern, so that specialized and expressive heterophilic GNN architectures can be derived.
Extensive experiments illustrate the superiority of Auto-HeG in developing excellent heterophilic GNNs to human-designed models and graph NAS models.
\end{abstract}
\maketitle
\section{Introduction}
Current mainstream research on graph neural networks (GNNs) principally builds upon the following two foundations: the homophily assumption of graph-structured data and the expertise dependency of GNN architecture design~\cite{tang2009social,peng2022reinforced,vashishth2019composition,wu2020connecting,ying2018graph,ma2019learning,zheng2022rethinking,zhang2022trustworthy}. 
At first, the homophily assumption defines that nodes with similar features or same class labels are linked together. 
For instance, research articles within the same research area would be linked more likely by citing each other.
In contrast, {\it{heterophily}}, the property that linked nodes have dissimilar features and different class labels, does not attract equal attention but widely exists in various web-related real-world applications~\cite{pandit2007netprobe,zhu2021graph,h2gcn_zhu2020beyond,zheng2022graph,liu2022beyond}.
For instance,
%, in online social networks related to dating, most people prefer to date people of the opposite gender~\cite{zhu2021graph}. 
in online transaction networks, fraudsters are more likely to build connections with customers instead of other fraudsters~\cite{pandit2007netprobe}.
At this point, heterophily of graphs plays an important role in web socio-economic system, and modeling the heterophily explicitly would benefit the development of web techniques and infrastructure.
Due to the differences of structure and property between homophilic and heterophilic graphs, existing GNNs, which target for modelling homophily, cannot be directly applied to heterophilic graphs, and the main reasons lie in the following two aspects~\cite{zheng2022graph}:
(1) local \vs. non-local neighbors: homophilic GNNs concentrate more on addressing local neighbor nodes within the same class, while on heterophilic graphs, informative same-class neighbors usually are non-local in graph topology; 
(2) uniform \vs. diverse aggregation: homophilic GNNs uniformly aggregate information from similar local neighbors and then accordingly update central (ego) nodes, while heterophilic graphs expect discriminative node representation learning to diversely extract information from similar and dissimilar neighbors.

Very recently, a few researchers divert their attention to develop heterophilic GNNs, typically by introducing higher-order neighbors~\cite{h2gcn_zhu2020beyond} or modelling homophily and heterophily separately in the message passing scheme~\cite{fagcn_bo2021beyond}.
Nevertheless, current heterophilic GNNs highly rely on the knowledge of experts to manually design models for graphs with different degrees of heterophily. 
For one thing, this process would cost many human efforts and the performance of derived GNNs would be limited by expertise. 
For another thing, real-world heterophilic graphs generally show the significant complexity and variety in terms of graph structure. 
It would be harder to artificially customize GNNs to make them adapt to various heterophilic graphs.

In light of this, automated graph neural network learning via neural architecture search (NAS)~\cite{liu2018darts,PDARTS2019progressive,PCNAS2020improving,NEAS2021one,AngleNAS2020angle,PAD-NASxia2022progressive,wang2021rethinking,zheng2022MRGNAS}, a line of research for saving human efforts in designing effective GNNs, would be a feasible way of tackling the dilemma of heterophilic GNN development.
Through learning in well-designed search spaces with efficient search strategies, automated GNNs via NAS have achieved promising progress on various graph data analysis tasks~\cite{gao2020graph,huan2021search,wang2021autogel,pan2021autostg,ding2021diffmg,zhang2020autosf}.
However, existing graph NAS studies, \eg, GraphNAS~\cite{gao2020graph} and SANE~\cite{huan2021search}, are all constrained under the homophily assumption.
To the best of our knowledge, there is still a research blank in developing heterophilic graph NAS for automated heterophilic graph learning.
The most straightforward way to implement heterophilic graph NAS might be replacing the homophilic operations in existing homophilic search spaces with heterophilic GNN operations, followed by universal gradient-based search strategies.
But this direct solution would incur some issues: 
First, it is difficult to determine what heterophilic operations are beneficial and whether certain homophilic operations should be kept for facilitating heterophilic graph learning, since the heterophilic graphs might be various and complex with different degrees of heterophily. 
Second, simply making existing universal search strategies adapt to new-defined heterophilic search spaces would limit the searching performance, resulting in suboptimal heterophilic GNN architectures. The heterophily should be integrated into the searching process for architecture optimization as well.

To tackle all the above-mentioned challenges of heterophilic graph NAS, in this paper, we propose a novel automated graph neural network on heterophilic graphs, namely Auto-HeG, to effectively learn heterophilic node representations via heterophily-aware graph neural architecture search. 
To the best of our knowledge, this is the first automated heterophilic graph learning method, and our theme is to explicitly incorporate heterophily into all stages of automatic heterophily graph learning, containing search space design, supernet training, and architecture selection. 
\textbf{For search space design:} 
By integrating joint micro-level and macro-level designs, Auto-HeG first builds a comprehensive heterophilic search space through the diverse message-passing scheme, enabling it to incorporate the complexity and diversity of heterophily better.
At the micro-level, Auto-HeG conducts non-local neighbor extension, ego-neighbor separation, and diverse message passing; At the macro-level, it introduces adaptive layer-wise combination operations.
\textbf{For supernet training:} 
To narrow the scope of candidate operations in the proposed heterophilic search space, Auto-HeG presents a progressive supernet training strategy to dynamically shrink the initial search space according to layer-wise variation of heterophily, resulting in a compact and efficient supernet.
\textbf{For architecture selection:}
Taking heterophily as the specific guidance, Auto-HeG derives a novel heterophily-aware distance as the criterion to select effective operations in the leave-one-out pattern, leading to specialized and expressive heterophilic GNN architectures. Extensive experiments on the node classification task illustrate the superior performance of the proposed Auto-HeG to human-designed models and graph NAS models.
In summary, our contributions are listed as follows:
\begin{itemize}[topsep=1.5pt,itemsep=2pt,leftmargin=15pt,parsep=1.5pt]
    \item We propose a novel automated graph neural network on heterophilic graphs by means of heterophily-aware graph neural architecture search, namely {\bf{Auto-HeG}}, to the best of our knowledge, for the first time.
    \item  To integrate the complex and various heterophily explicitly, we build a comprehensive heterophilic GNN search space that incorporates non-local neighbor extension, ego-neighbor separation, diverse message passing, and layer-wise combination at the micro-level and the macro-level.
    \item To learn a compact and effective heterophilic supernet, we introduce a progressive supernet training strategy to dynamically shrink the initial search space, enabling the narrowed searching scope with layer-wise heterophily variation.
    \item To select optimal GNN architectures specifically instructed by heterophily, we derive a heterophily-aware distance criterion to develop powerful heterophilic GNNs in the leave-one-out manner and extensive experiments verify the superiority of Auto-HeG.
\end{itemize}
\section{Related work}\label{related_work}
%\vspace{-0.3cm}
%\subsection{}
%\noindent\textbf{Graph Neural Networks with Heterophily.}
\paragraph{Graph Neural Networks with Heterophily.}
Existing heterophilic GNNs mainly work on two aspects~\cite{zheng2022graph}: non-local neighbor extension~\cite{abu2019mixhop,h2gcn_zhu2020beyond,jin2021universal,geom_pei2020geom,wang2021powerful,bm_gcn_he2021block} and GNN architecture refinement~\cite{fagcn_bo2021beyond,acm_luan2021heterophily,yang2021diverse,yan2021two,SureshBNLM21,chen2020simple,chien2020adaptive}.
In detail, non-local neighbor extension methods focus on exploring the informative neighbor set beyond local topology. 
%In this way, the neighbors that share the same labels with the central node could be captured, even if they are located in the long-range non-local topology. 
In contrast, GNN architecture refinement methods design heterophily-specific message-passing models to learn discriminative node representations. 
%By distinctly aggregating different information according to the different degrees of heterophily, heterophilic GNN architectures could be designed based on the diverse message-passing scheme.
Typically, Mixhop~\cite{abu2019mixhop} and H2GCN~\cite{h2gcn_zhu2020beyond} introduced higher-order neighbor mixing of local one-hop neighbors and non-local $K$-hop neighbors to learn discriminative central node representations. 
And Geom-GCN~\cite{geom_pei2020geom} defined the split 2D Euclidean geometry locations as the geometric relationships with different degrees of heterophily, enabling it to discover potential neighbors for further effective aggregation.
In contrast, FAGCN~\cite{fagcn_bo2021beyond} developed diverse aggregation functions by introducing low-pass and high-pass filters, corresponding to learning homophily and heterophily, respectively.
Besides, GCNII~\cite{chen2020simple} and GPR-GNN~\cite{chien2020adaptive} considered the layer-wise representation integration to boost the performance of GNNs with heterophily.
Despite the promising development, these heterophilic GNNs highly rely on the expertise to manually design GNNs for tackling the complex and diverse heterophily. 
For one thing, it would cost exhausted human efforts. 
For another thing, the design scopes and flexibility would be constrained by expert knowledge, leading to limited model performance. 
In light of this, we are the first to propose an automated graph neural architecture search framework for heterophilic graphs, to significantly save human efforts and automatically derive powerful heterophilic GNN architectures with excellent learning abilities.
\paragraph{Graph Neural Architecture Search.}
Graph NAS has greatly enlarged the design picture of automated GNN development for discovering excellent and powerful models~\cite{gao2020graph,zhou2019auto,SNAG20ZhaoWY,huan2021search, wang2021autogel,pan2021autostg,ding2021diffmg,zhang2020autosf}.
Generally, the development of graph NAS methods focuses on two crucial research aspects: search space and search strategy. 
The former defines architecture and functional operation candidates in a set space, and the latter explores the powerful model components in the defined search space.
Typically, GraphNAS~\cite{gao2020graph} and AGNN~\cite{zhou2019auto} constructed the micro-level search space containing classical GNN components and related hyper-parameters, followed by architecture controllers based reinforcement learning (RL) search strategy.
And SNAG~\cite{SNAG20ZhaoWY} further simplified GraphNAS at the micro-level and introduced the macro-level inter-layer architecture connections.
Based on this search space, SANE~\cite{huan2021search} implemented DARTS~\cite{liu2018darts}, a gradient-based search strategy, to automatically derive effective GNN architectures on graphs in a differential way.
Nevertheless, current graph NAS methods are still constrained under the homophily assumption and cannot learn explicitly and effectively on graphs with complex and diverse heterophily.
Therefore, we draw inspiration from the lines of NAS research on search space shrinking~\cite{PDARTS2019progressive,PCNAS2020improving,NEAS2021one,AngleNAS2020angle,PAD-NASxia2022progressive}, supernet optimization~\cite{SNAS19,4gpu2019searching}, and architecture selection~\cite{wang2021rethinking}, and importantly extend them to the development of automated heterophilic GNNs.
Our critical goal is to build a well-designed search space with a customized search strategy via graph NAS, to explicitly integrate the heterophily into full stages of automated heterophilic GNN development.
\begin{figure*}[t]
    \centering
    \includegraphics[width=0.65\textwidth]{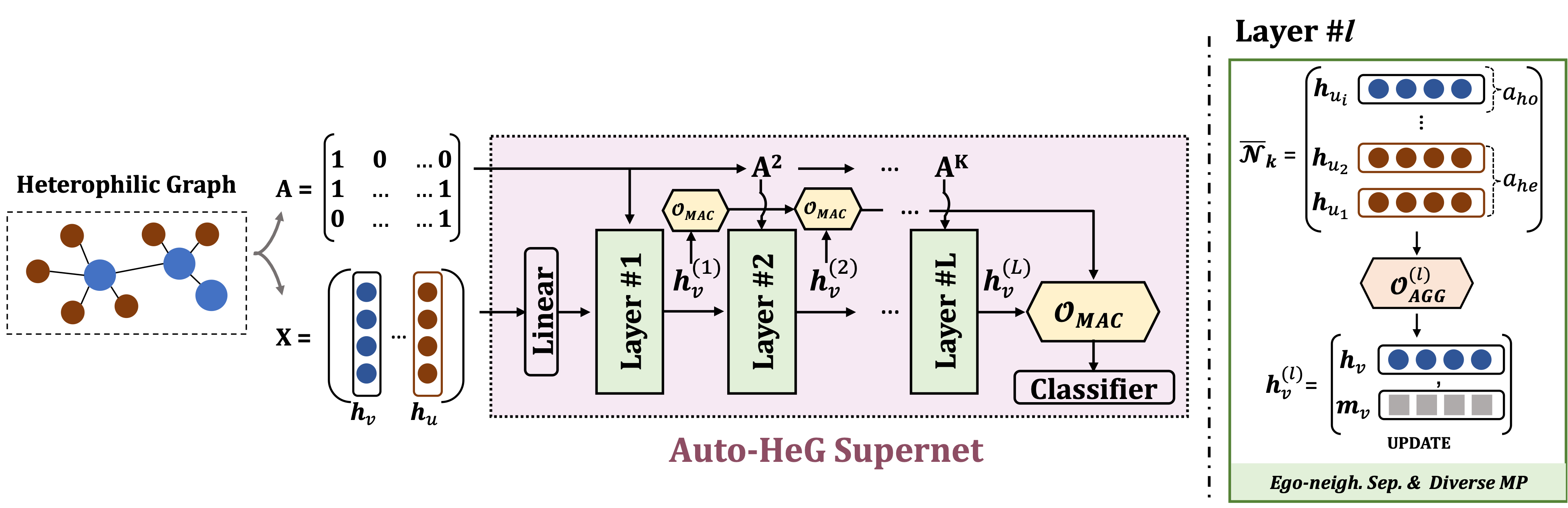}
    \vspace{-10pt}
    \caption{Overall supernet of the proposed Auto-HeG. \scriptsize{{$\mathbf{A}$ denotes the adjacency matrix and $\mathbf{A}^{k}$ denotes the $k$-th hop neighbors for $k=\{1,2,\cdots,K\}$, and the right part shows the details of $l$-th layer with the layer-wise heterophily-specific operation space $\mathcal{O}_{AGG}^{(l)}$ via progressive supernet training.}}}
    \label{fig:frame}
    \vspace{-8pt}
\end{figure*}
\section{Auto-HeG: Automated graph neural network on heterophilic graphs}\label{method}
%\vspace{-0.2cm}
\subsection{Preliminary.}
%\vspace{-0.1cm}
%\noindent\textbf{Uniform Message Passing.} 
\paragraph{Uniform Message Passing.}
Let $\mathcal{G} = (\mathcal{V}, \mathcal{E})$ be an undirected, unweighted graph where $\mathcal{V} = \{v_1, \cdots, v_{|\mathcal{V}|} \}$ is the node set and $\mathcal{E} \in \mathcal{V} \times \mathcal{V}$ is the edge set. The neighbor set of node $v$ is $\mathcal{N}(v) = \{u: (v, u) \in \mathcal{E}  \}$, and initial node features are represented by ${\mathbf{X}} \in \mathbb{R}^{|\mathcal{V}| \times {d_{0}}}$ with $d_{0}$-dimension features. For the uniform message passing scheme under the homophily assumption, node representations of GNNs are learned by first aggregating the messages from local neighbors, and then updating the ego-node representations by combining the aggregated messages with themselves~\cite{gin_xu2019powerful}. This process can be denoted as:
\begin{equation}
\label{eq:GNN}
\begin{aligned}
\mathbf{m}^{(l)}_v  &= \operatorname{AGG}^{(l)}(\{\mathbf{h}^{(l-1)}_u: u \in \mathcal{N}(v)\}),\\
% \mathbf{m}^{(l)}_v & = \operatorname{AGGREGATE}^{(l)}(\{a_{uv}^{(l)}\,\mathbf{h}^{(l-1)}_u: u \in \mathcal{N}(v)\}),
\mathbf{h}^{(l)}_{v}  &=\operatorname{UPDATE}^{(l)}(\mathbf{h}^{(l-1)}_v, \mathbf{m}^{(l)}_{v}),
\end{aligned}
\end{equation}

\noindent where $\mathbf{m}^{(l)}_v$ and $\mathbf{h}^{(l)}_{v}$ are the message vector and the representation vector of node $v$ at the $l$-th layer, respectively. $\operatorname{AGG}(\cdot)$ and $\operatorname{UPDATE}(\cdot)$ are the aggregation function and update function, respectively.
Given the input of the first layer, the learned node representations with ${d_{1}}$ dimensions at each layer of $L$-layer GNN can be denoted as $\mathbf{H}^{(l)} \in \mathbb{R}^{|\mathcal{V}| \times {d_{1}}}$ for $ l=\{1,2,\cdots,L\}$.
%Given the input of the first layer $\mathbf{H}^{(0)} = \mathbf{X}\in \mathbb{R}^{|\mathcal{V}| \times {d_{0}}}$, the learned node representations with ${d_{1}}$ dimensions at each layer of $L$-layer GNN can be denoted as $\mathbf{H}^{(l)} \in \mathbb{R}^{|\mathcal{V}| \times {d_{1}}}$ for $ l=\{1,2,\cdots,L\}$.
%Given the input of the first layer as $\mathbf{H}^{(0)} = \mathbf{X}$, the learned node representations at each layer of $L$-layer GNN can be denoted as $\mathbf{H}^{(l)} \in \mathbb{R}^{|\mathcal{V}| \times {d_{1}}}$ for $ l=\{1,2,\cdots,L\}$.

%\noindent\textbf{Measure of Heterophily \& %Homophily.} 
\paragraph{Measure of Heterophily \& Homophily.}
In general, heterophily and homophily of a graph $\mathcal{G} = (\mathcal{V}, \mathcal{E})$ can be measured by node homophily~\cite{geom_pei2020geom}. Concretely, the node homophily $\gamma_{node}$ is the average proportion of the neighbors with the same class of each node as: $\gamma_{node} = 1/{|\mathcal{V}|} \sum_{v \in \mathcal{V}} (|\{u \in \mathcal{N}(v): y_v = y_u \}|/{|\mathcal{N}(v)|}).$
%\begin{equation}
%\label{eq:node_homo}
%\gamma_{node} = \frac{1}{|\mathcal{V}|} \sum_{v \in \mathcal{V}} \frac{|\{u \in \mathcal{N}(v): y_v = %y_u \}|}{|\mathcal{N}(v)|}.
%\end{equation}

The range of $\gamma_{node}$ is $[0,1]$. Graphs with strong homophily have higher $\gamma_{node}$ (closer to $1$); Whereas graphs with strong heterophily have smaller $\gamma_{node}$ (closer to $0$).

\subsection{Heterophilic Search Space Design}
To incorporate the complexity and variety of heterophily on graphs, we design a comprehensive heterophilic search space in the proposed Auto-HeG, involving joint micro-level and macro-level candidate operations. 
At the micro-level, the proposed search space contains three important components: non-local neighbor extension, ego-neighbor separation, and diverse message passing; 
While the macro-level consists of intermediate layer combination functions and skip connections for integrating layer-wise node representations.
The well-designed search space is the basis for building a supernet for search effective architectures on heterophilic graphs as shown in Fig.~\ref{fig:frame}, and we will discuss this further in Sec.~\ref{sec:supernet}. 
In the following, we mainly give detailed descriptions of the candidate operations in the proposed heterophilic search space, where its summary is listed in Table~\ref{tab:operations}. 
%Based on the well-designed search space, we build the overall Auto-HeG supernet as shown in Fig.~\ref{fig:frame}.
%The overall Auto-HeG supernet built based on the proposed heterophilic search space is shown in Fig.~\ref{fig:frame}, and more search space details are listed in Table~\ref{tab:operations}.

%\vspace{-0.2cm}
\subsubsection{Micro-level Design}
%\vspace{-0.1cm}
%\noindent\textbf{Non-local Neighbor Extension.} 
\paragraph{Non-local Neighbor Extension.}
Homophilic graphs generally take local nodes from one hop away as neighbors of the ego node for message aggregation, since the same-class nodes mainly locate in their proximal topology.
On the contrary, on heterophilic graphs, nodes within the same class could be far away from each other. 
That means only considering the one-hop-away neighbors would be insufficient to capture the heterophily on graphs explicitly.
Hence, to break the local topology limitation under the homophily assumption, we extend local one-hop neighbors to non-local $K$-hop neighbors to incorporate heterophilic neighbor information.
Concretely, we modify the uniform message passing scheme in Eq.~(\ref{eq:GNN}) in terms of $K$-hop neighbors as follows:
%Different from homophilic graphs mainly taking local nodes from one-hop away as neighbors for message aggregation, nodes within the same class on heterophilic graphs could be far away from each other. 
%To break such local topology limitation, we extend local one-hop neighbors to non-local $K$-hop neighbors by modifying the uniform message passing scheme in Eq.~\ref{eq:GNN} as follows:
\begin{equation}
\label{eq:neighbor}
\mathbf{m}^{(l)}_v = \operatorname{AGG}^{(l)}(\{\mathbf{h}^{(l-1)}_u: u \in \mathcal{N}_{k}(v)\}), \, k=\{1,2,\cdots,K\},
\end{equation}
where $\mathcal{N}_{k}(v)$ denotes the $k$-th hop neighbor set of node $v$. 
In this way, the derived heterophilic GNNs could mix latent information from neighbors within the same class at various distances of graph topology.
Specifically, to avoid multi-hop neighbors bringing the exponential explosion of graph scale, we restrict the $k$-hop neighbor set as neighbors that are connected by at least $k$ different paths to the ego nodes. 
For instance, a two-hop neighbor set contains neighbors that are two hops away from ego nodes and have at least two different paths to reach the ego nodes.
Note that, different from existing heterophilic GNNs (\eg, Mixhop~\cite{abu2019mixhop} and H2GCN~\cite{h2gcn_zhu2020beyond}) introducing multi-hop neighbors at each layer and combining them in parallel, we relate the number of neighbor hops to the number of GNN layers correspondingly in the proposed Auto-HeG.
The intuition behind this is to alleviate the complexity of the search space but keep its effectiveness at the same time.
%The intuition behind this is to alleviate the complexity of the search space but keep its effectiveness at the same time.
%Note that, different from existing heterophilic GNNs (\eg, Mixhop~\cite{abu2019mixhop} and H2GCN~\cite{h2gcn_zhu2020beyond}) introducing multi-hop neighbors at each layer and combining them in parallel, we relate the number of neighbor hops to the index of GNN layers correspondingly in the proposed Auto-HeG. 
%The intuition behind this is to alleviate the complexity of the search space but keep its effectiveness at the same time.

%\begin{wraptable}{r}{8.5cm}
%\vspace{-10pt}
\begin{table}[b]
\vspace{-15pt}
\caption{Heterophilic search space details of the proposed Auto-HeG. {\scriptsize{`homo.' and `hete.' indicate homophily-related and heterophily-related aggregation functions, respectively.}}}
\vspace{-5pt}
\label{tab:operations}
\centering
%\large
%\setlength{\tabcolsep}{2.5pt}
\resizebox{0.45\textwidth}{!}{
\begin{tabular}{l|ll|l}
\toprule
Search Space                 & \multicolumn{2}{l|}{Modules}                        & Operations                                                                                                                                                 \\ \midrule
\multirow{3}{*}{Micro-level} & \multicolumn{2}{l|}{Neighbors}                     & $\{A, A^{2}, \cdots, A^{K}\}$                                                                                                                                                  \\ \cmidrule(r){2-4} %\specialrule{0em}{2pt}{2pt}
                             & \multicolumn{1}{l|}{\multirow{2}{*}{$\mathcal{O}_{AGG}$}} & {homo.}  & \makecell[l]{\small{\{	
\texttt{SAGE,\,SAGE\_SUM,\,SAGE\_MAX,GCN,\,GIN,\,GAT,\,GAT\_SYM,}} \\  \small{	
\texttt{\,GAT\_COS,\,GAT\_LIN,\,GAT\_GEN\_LIN,\,GeniePATH}\}}} \\ \cmidrule(r){3-4} %\specialrule{0em}{2pt}{2pt}
                             & \multicolumn{1}{l|}{}                      & {hete.} & {\small{\{	
\texttt{GCNII,\,FAGCN,\,GPRGNN,\,SUPERGAT,\,GCN\_CHEB,\,APPNP,\,SGC}\}}}                                                                                 \\ \midrule
Macro-level                  & \multicolumn{2}{l|}{$\mathcal{O}_{MAC}$}                          & $l\_skip$, $l\_zero$, $l\_concat$, $l\_max$, $l\_lstm$ \\ \bottomrule
\end{tabular}
}
\vspace{-8pt}
%\end{wraptable}
%\vspace{-0.15cm}
\end{table}

%\noindent\textbf{Ego-neighbor Separation.} 
\paragraph{Ego-neighbor Separation.}
Despite capturing the messages from non-local neighbors, some local neighbors still have dissimilar class labels with the ego nodes.
Hence, it is necessary to separate the ego node representations from their neighbor representations, as verified by the work~\cite{h2gcn_zhu2020beyond}.
In this way, heterophilic GNNs could learn the ego node features by discriminatively combining the aggregated neighbor information with themselves at the following update step.
%learn feature representations of ego and neighbor nodes, as verified by the work~\cite{h2gcn_zhu2020beyond}.
%combining the information of ego nodes and their neighbor nodes is another key to learning discriminative representations on graphs with heterophily.
%Considering the ego nodes and their local neighbor nodes usually are dissimilar from class labels on heterophilic graphs, it is necessary to separately learn feature representations of ego and neighbor nodes, as verified by the work~\cite{h2gcn_zhu2020beyond}.
At this point, we further modify the uniform message passing scheme at the aggregation step in Eq.~(\ref{eq:neighbor}) as
%At this point, we go further step to modify the uniform message passing scheme in Eq.~\ref{eq:neighbor} as
\begin{equation}
\label{eq:ego}
\mathbf{m}^{(l)}_v = \operatorname{AGG}^{(l)}(\{\mathbf{h}^{(l-1)}_u: u \in \bar{\mathcal{N}}_{k}(v)\}), \, k=\{1,2,\cdots,K\},
\end{equation}
where $\bar{\mathcal{N}}_{k}(v)$ denotes the neighbor set that excludes the ego node $v$, \ie, removing self-loops in the original graph structure.

%\noindent\textbf{Diverse Message Passing.} 
\paragraph{Diverse Message Passing.}
Given the extended non-local neighbor set, diverse message-passing scheme is the core of tackling the heterophily on graphs. 
By distinguishing the information of similar neighbors (likely in the same class) from that of dissimilar neighbors (likely in different classes), heterophilic GNNs could aggregate discriminative messages from diverse neighbors.
Typically, we decompose the edge-aware weights $\mathbf{a}_{uv}$ into homophily-component $a_{uv}^{ho}$ and heterophily-component $a_{uv}^{he}$ on different neighbor representations $\mathbf{h}_{u}$, respectively. 
Hence, we can obtain $\mathbf{a}_{uv}=[a_{uv}^{ho};\,a_{uv}^{he}]$.
Moreover, due to the complexity and variety of heterophily, there might be no adequate prior knowledge of the extent of heterophily on graphs. 
Certain homophilic aggregation functions, \eg, GAT~\cite{velivckovic2018graph}, also have the ability to impose discriminative weights on different neighbors. 
Hence, we first introduce ample heterophilic and homophilic aggregation functions for the comprehensiveness of the initial search space design. 
Adaptively learning a more compact search space is postponed to the latter supernet training stage. 
Specifically, we introduce 18 homophilic and heterophilic aggregation functions, denoting as the diverse aggregation function set $\mathcal{O}_{AGG}$ shown in Table~\ref{tab:operations}. 
In this way, the heterophilic message-passing scheme can be denoted as:
%Specifically, we introduce 18 aggregation functions based on prevalent homophilic and heterophilic GNNs. The diverse aggregation function set $\mathcal{O}_{AGG}$ is presented in Table~\xin{tab}. At this point, we further extend the heterophilic message passing as:
\begin{equation}
\label{eq:diverseMP}
\mathbf{m}^{(l)}_v = \mathcal{O}_{AGG}^{(l)}(\{\mathbf{a}_{uv}^{l-1}\mathbf{h}^{(l-1)}_u: u \in \bar{\mathcal{N}}_{k}(v)\}), \, k=\{1,2,...,K\}.
\end{equation}

%\vspace{-0.5cm}
In summary, the micro-level design of the proposed heterophilic search space mainly works on diverse message passing with extended local and non-local neighbors, along with separated ego and neighbor node representation learning. 
And such design encourages heterophilic GNNs to adaptively incorporate the heterophily of graphs in each phase of message passing.

\subsubsection{Macro-level Design}
%\paragraph{Layer-wise Operations}
To integrate local and global information from different layers, combining intermediate layer representations and introducing skip connections have been verified as beneficial layer-wise operations in human-designed GNNs~\cite{li2019deepgcns,xu2018representation,h2gcn_zhu2020beyond} and automated GNNs~\cite{huan2021search,you2020design,wang2021autogel}. 
In light of this, to enable the flexible GNN architecture design on heterophilic graphs, we introduce the inter-layer connections and combinations into the macro-level search space.
In this way, the automated heterophilic GNNs could capture hierarchical information at different layers of network architectures. 
%Hence, we introduce the inter-layer connections and combinations into the macro-level search space, leading to flexible heterophilic GNN architectures. 
In detail, our macro-level space contains 5 candidate operations as  ${\mathcal{O}}_{MAC} = \{l\_skip, l\_zero, l\_concat, l\_max, l\_lstm\}$. 
So that the final output of a heterophilic GNN can be denoted as:
%Combining intermediate layer representations and introducing skip connections, have been verified as beneficial layer-wise operations for integrating local and global information from different layers in human-designed GNNs~\cite{li2019deepgcns,xu2018representation,h2gcn_zhu2020beyond} and automated GNNs~\cite{huan2021search,you2020design,wang2021autogel}. To develop flexible heterophilic GNN architectures, we introduce the inter-layer connections and combinations into the macro-level search space. In detail, our macro-level space contains 5 candidate operations as  ${\mathcal{O}}_{MAC} = \{l\_skip, l\_zero, l\_concat, l\_max, l\_lstm\}$, so that the final output can be denoted as:
\begin{equation}
\label{eq:macro}
\mathbf{h}_{out} = \mathcal{O}_{MAC}[\mathbf{h}^{(1)}, \mathbf{h}^{(2)}, \cdots, \mathbf{h}^{(L)}].
\end{equation}

With the joint micro-level and macro-level candidate operations, the proposed heterophilic search space significantly enlarges the design scopes and the flexibility in developing GNN architectures on graphs with heterophily.
\subsection{Progressive Heterophilic Supernet Training}\label{sec:supernet}
%\subsection{Progressive Supernet Training on Heterophilic Search Space}
Based on the proposed heterophilic search space, we build a one-shot heterophilic supernet as shown in Fig.~\ref{fig:frame}. 
Given the heterophilic graph with input feature $\mathbf{X}\in \mathbb{R}^{|\mathcal{V}| \times {d_{0}}}$ and adjacency matrix $\mathbf{A}\in \mathbb{R}^{|\mathcal{V}| \times |\mathcal{V}|}$, the proposed Auto-HeG supernet first implements a linear layer, \ie, Multi-layer Perceptron (MLP), to obtain the initial node embedding $\mathbf{H}^{(0)} \in \mathbb{R}^{|\mathcal{V}| \times {d_{1}}}$. 
Then, the node embedding $\mathbf{H}^{(0)}$ and the adjacency matrix $\mathbf{A}$ would be fed into the first layer containing micro-level candidate operations with the $\mathcal{O}_{AGG}$ set, leading to the output node representation $\mathbf{H}^{(1)}$.
Next, $\mathbf{H}^{(1)}$ is fed into the macro-level $\mathcal{O}_{MAC}$ for the layer-wise connection. 
Meanwhile, $\mathbf{H}^{(1)}$ and the two-hop adjacency matrix $\mathbf{A}^{2}$ would be the inputs of the next micro-level layer.
The above process implements repeatedly layer by layer before the final node representations experience the classifier to obtain the predictions of node classes.

Even though the proposed search space contains as many candidate operations as possible, real-world heterophilic graphs usually show significant differences in complexity and diversity, and we still lack prior knowledge on what heterophilic and homophilic aggregations are beneficial for heterophilic graph learning. 
The proposed initial search space might be comprehensive but severely challenge the effectiveness of supernet training to automatically derive expressive GNN architectures, especially when heterophily varies on different graphs.
In light of this, a possible solution comes: why not let the proposed supernet adaptively keep the beneficial operations and drop the irrelevant counterparts as the process of iterative learning goes on? 
This would contribute to a more compact supernet with relevant candidate operations specifically driven by current heterophilic graphs and tasks, leading to more effective heterophilic GNN architecture development.
Moreover, considering some operations will never be selected by certain layers in the final architectures mentioned by~\cite{PAD-NASxia2022progressive}, this solution ensures different layers flexibly customize their own layer-wise candidate operation sets according to heterophily variation, rather than sharing the entire fixed and large search space with all layers.
\begin{algorithm}[t]
\caption{Progressive Heterophilic Supernet Training}  
\label{alg:A}  
\begin{algorithmic}[1]
\REQUIRE {Initial heterophilic supernet $\mathcal{S}_{0}$, number of shrinking iterations $T$, number of candidate operations $C$ to be dropped per iteration $t$.}
\ENSURE {Compact heterophilic supernet $\mathcal{S}_{c}$.}
\STATE {Let $\mathcal{S}_{c} \gets \mathcal{S}_{0}$;}
\WHILE {$t<T$}
\STATE {Training  $\mathcal{S}_{c}$ for several epochs as Eq.~(\ref{eq:bi_opt}) and~(\ref{eq:o_ij});}
\STATE {Ranking the magnitudes of the architecture  ${\bm{\alpha}}$;}
\STATE {Dropping $C$ operations from $\mathcal{S}_{c}$ with the smallest $C$ architecture weights;}
\ENDWHILE
\end{algorithmic}  
\end{algorithm} 

Therefore, to derive a more compact heterophilic one-shot supernet for efficient architecture design, we present a progressive training strategy to dynamically shrink the initial search space and adaptively design the candidate operation space in each layer. 
Specifically, let $\mathcal{S}_{0} =(\mathcal{E}^{\mathcal{S}}_{0}, \mathcal{N}^{\mathcal{S}}_{0}) $ denotes the initial heterophilic supernet constructed based on the proposed entire search space $\mathcal{O}$, and $\mathcal{E}^{\mathcal{S}}_{0}$ and $\mathcal{N}^{\mathcal{S}}_{0}$ are the edge set and the node set, respectively. 
We first train $\mathcal{S}_{0}(\bm{\alpha},\bm{w})$ several steps for stochastic differentiable search based on the bi-level optimization:
\begin{equation}
\begin{small}
\label{eq:bi_opt}
\begin{aligned}
    \min _{\bm{\alpha}} &\quad \mathcal{L}_{ {val }}\left(\bm{w}^{*}(\bm{\alpha}), \bm{\alpha}\right),\\
    s.t. &\quad \bm{w}^{*}(\bm{\alpha})=\operatorname{argmin}_{\bm{w}} \mathcal{L}_{{train }}(\bm{w}, \bm{\alpha}),
\end{aligned}
\end{small}
\end{equation} 
which conducts iterative learning of the architecture weight $\bm{\alpha}$ and the model parameter $\bm{w}$. 
Let $o \in \mathcal{O}$ denotes a certain operation in heterophilic search space $\mathcal{O}$, and a node pair $(\mathbf{x}_{i},\mathbf{x}_{j})$ denotes the latent vectors from $i$-th node to $j$-th node of the supernet, the learned architecture weights ${\bm{\alpha}} =\left\{\alpha_{o}^{(i, j)}| o \in \mathcal{O}, (i,j) \in \mathcal{N}^{\mathcal{S}}_{0} \right\}$, and the operation specific weight vector $\alpha_{o}^{(i, j)}$ can be derived as:
\begin{equation}
\begin{small}
\label{eq:o_ij}
\bar{o}^{(i, j)}(\mathbf{x})=\sum_{o \in \mathcal{O}} \frac{\exp \left[\left(\log \alpha_{o}^{(i, j)} + u_{o}^{(i, j)} \right)/ \tau\right]}{\sum_{o^{\prime} \in \mathcal{O}} \exp \left[\left(\log \alpha_{o^{\prime}}^{(i, j)}+ u_{o^{\prime}}^{(i, j)}\right) / \tau\right]}\, \cdot o(\mathbf{x})
\end{small}
\end{equation}
where $\bar{o}^{(i, j)}$ is the mixed operation of edges between $(i,j)$, and $u_{o}^{(i, j)} = -\log \left(-\log \left(U\right)\right)$ where $U\sim \operatorname{Uniform}(0, 1)$. $\tau$ is the temperature factor that controls the extent of the continuous relaxation. When $\tau$ is closer to 0, the weights would be closer to discrete one-hot values.

To further exploit the relevance and importance of candidate operations to heterophily, we rank the obtained ${\bm{\alpha}}$ layer by layer based on its magnitudes and drop the last $C$ irrelevant operations, \ie, cutting $C$ number of edges in each layer at the current supernet training stage. 
By repeating the above process $T$ iterations, we gradually and dynamically shrink the initial search space and compress the supernet, leading to a compact and effective supernet $\mathcal{S}_{c}$ by customizing layer-wise heterophilic candidate operation set. The overall process of the proposed progressive supernet training strategy is summarized in Algo.~\ref{alg:A}.
\begin{algorithm}[t]
\caption{Heterophily Guided Architecture Selection}
\label{alg:B}
\begin{algorithmic}[1]
\REQUIRE {Pretrained heterophilic supernet $\mathcal{S}_{c}$ from Algo.~\ref{alg:A},
edge set $\mathcal{E}^{\mathcal{S}}_{c}$ and node set $\mathcal{N}^{\mathcal{S}}_{c}$ of $\mathcal{S}_{c}$.}
\ENSURE {A heterophilic GNN model with the set of selected operations $\{o^{*}_{e}|{e\in\mathcal{E}^{\mathcal{S}}_{c}}\}$.}
\WHILE {$|\mathcal{E}^{\mathcal{S}}_{c}|>0$}
\STATE {randomly remove an edge ${e_{i}\in\mathcal{E}^{\mathcal{S}}_{c}}$ from $\mathcal{E}^{\mathcal{S}}_{c}$;}
\FORALL{operations $o \in \mathcal{O}$ on edge $e_{i}$}
\STATE {calculate $D_{hete}{(\backslash o)}$ when $o$ is removed;} 
\ENDFOR
\STATE {select the best operation for $e_{i}$: $o^{*}_{e} \gets \arg \min_{o} D_{hete}(\backslash o)$;}
\ENDWHILE
\end{algorithmic}
\end{algorithm}

\subsection{Heterophily-guided Architecture Selection}
%\vspace{-0.1cm}
With the magnitudes of the architecture weights as the metric, Auto-HeG progressively learns a compact supernet with layer-wise search spaces driven by the heterophily variation. 
Considering an extreme case: the supernet shrinks gradually till every edge of it keeps only one operation. Basically, that is the general strategy used by current graph NAS methods, \ie, argmax-based architecture selection scheme.
That means only the operations corresponding to the maximum edge weights in the architecture supernet could be preserved by such an architecture selection scheme to build the ultimate GNNs.
%However, very recent research~\cite{wang2021rethinking} has verified that selecting the final model architectures resorting to architecture weight magnitudes might not be effective enough to indicate the operation strength.
However, very recent research~\cite{wang2021rethinking} has verified the architecture magnitude is not effective enough to indicate the operation strength in the final GNN architecture selection.
That is why we only use this strategy as the beginning to generally narrow the scope of candidate operations in the proposed search space.

%Therefore, to make the architecture selection of the compact supernet adapt to heterophily, we derives a heterophily-aware distance as the criterion to guide architecture selection for heterophilic graph learning.
Therefore, to select powerful GNN architectures specifically instructed by heterophily, we derive a heterophily-aware distance as the criterion to guide heterophilic GNN architecture selection.
And inspired by the perturbation-based architecture selection scheme in~\cite{wang2021rethinking}, we implement a leave-one-out manner to directly evaluate the contribution of each candidate operation to the compact supernet performance. 
Specifically, the proposed heterophily-aware distance $D_{hete}$ can be defined with the Euclidean distance as $D_{hete} = ||\hat{\mathcal{H}}-\mathcal{H}||^2$,
%\begin{equation}
%    D_{hete} = ||\hat{\mathcal{H}}-\mathcal{H}||^2,
%\end{equation}
where $\mathcal{H}$ and $\hat{\mathcal{H}}$ are the heterophilic matrices denoted as:
\begin{equation}
\begin{aligned}
    \mathcal{H} =\left(Y^{T} A Y\right) \oslash\left(Y^{T} A E\right), \, \hat{\mathcal{H}} =\left(\hat{Y}^{T} A \hat{Y}\right) \oslash\left(\hat{Y}^{T} A E\right),
\end{aligned}
\end{equation}
where $Y \in \mathbb{R}^{|\mathcal{V}| \times {p}}$ is the ground-truth label matrix of the heterophilic graph with $p$ classes of nodes, $E\in \mathbb{R}^{|\mathcal{V}| \times {p}}$ is an all-ones matrix, and $\oslash$ denotes the Hadamard division operation. Concretely, $\mathcal{H}_{i,j}$ indicates the connection probability of nodes \cite{bm_gcn_he2021block,zhu2021graph} between the $i$-th class and the $j$-th class with the ground truth labels, and $\hat{\mathcal{H}}_{i,j}$ works in the same way but with the predicted labels $\hat{Y}\in \mathbb{R}^{|\mathcal{V}| \times {p}}$, \ie, the output of the compact heterophilic supernet. 

The proposed heterophily-aware distance $D_{hete}$ explicitly restricts the connection possibility of nodes in any two classes predicted by the supernet, to be close to that of the ground truth. 
And smaller $D_{hete}$ indicates the better discriminative ability of a certain candidate operation to heterophilic node classes.
Taking $D_{hete}$ as the guidance, we select the optimal heterophilic GNN architecture from the pretrained compact supernet $\mathcal{S}_{c}$, and the whole process is illustrated in Algo.~\ref{alg:B}.
\begin{table*}[t]
\centering
\caption{Performance (ACC\%${\scriptstyle\pm}$std) of the proposed Auto-HeG compared with human-designed and graph NAS models on high-heterophily datasets. {\scriptsize{The best results are in bold and the second-best results are underline. Superscript $^{*}$ represents the officially reported results with the same dataset splits, where Geom-GCN and GCNII do not provide the std; And the remains are our reproduced results if official methods do not test under the same dataset splits.}}}
\vspace{-5pt}
\label{tab:exp_hete}
\resizebox{0.85\textwidth}{!}{
\begin{tabular}{l|l|p{2.5cm}<{\centering}p{2.5cm}<{\centering}p{2.5cm}<{\centering}p{2.5cm}<{\centering}}
\toprule
Methods                                                                                 & Datasets      & Cornell      & Texas         & Wisconsin    & Actor        \\\midrule
\multirow{9}{*}{\begin{tabular}[c]{@{}l@{}}Human-designed models\end{tabular}} & H2GCN-1$^{*}$   & $\underline{82.16{\scriptstyle\pm4.80}}$ & $\underline{84.86 {\scriptstyle\pm6.77}}$  & $\underline{86.67 {\scriptstyle\pm4.69}}$ & $\underline{35.86{\scriptstyle\pm1.03}}$ \\
                                                                                        & H2GCN-2$^{*}$   & $82.16 {\scriptstyle\pm6.00}$ & $82.16{\scriptstyle\pm5.28}$  & $85.88{\scriptstyle\pm4.22}$ & $35.62{\scriptstyle\pm1.30}$ \\
                                                                                        & MixHop$^{*}$        & $73.51{\scriptstyle\pm6.34}$ & $77.84{\scriptstyle\pm7.73}$  & $75.88{\scriptstyle\pm4.90}$ & $32.22{\scriptstyle\pm2.34}$ \\
                                                                                        & GPR-GNN       & $81.89{\scriptstyle\pm5.93}$ & $83.24{\scriptstyle\pm4.95}$ & $84.12{\scriptstyle\pm3.45}$ & $35.27{\scriptstyle\pm1.04}$ \\
                                                                                        & GCNII$^{*}$         & $76.49$        & $77.84$         & $81.57$        & - \\
                                                                                        & Geom-GCN-I$^{*}$    & $56.76$        & $57.58$         & $58.24$        & $29.09$        \\
                                                                                        & Geom-GCN-P$^{*}$    & $60.81$        & $67.57$         & $64.12$        & $31.63$        \\
                                                                                        & Geom-GCN-S$^{*}$    & $55.68$        & $59.73$         & $56.67$        & $30.30$
                                                                                    \\
                                                                                        & FAGCN         & $81.35{\scriptstyle\pm5.05}$	& $84.32{\scriptstyle\pm6.02}$	& $83.33{\scriptstyle\pm2.01}$	& $35.74{\scriptstyle\pm0.62}$
                                                                                        \\\midrule
\multirow{5}{*}{Graph NAS models}                                                      & GraphNAS      & $58.11{\scriptstyle\pm3.87}$  & $54.86{\scriptstyle\pm6.98}$  & $56.67{\scriptstyle\pm2.99}$ & $25.47{\scriptstyle\pm1.32}$ \\
                                                                                        & SNAG          & $57.03{\scriptstyle\pm3.48}$ & $62.70{\scriptstyle\pm5.52}$  & $62.16{\scriptstyle\pm4.63}$ & $27.84{\scriptstyle\pm1.29}$ \\
                                                                                        & SANE          & $56.76{\scriptstyle\pm6.51}$ & $66.22{\scriptstyle\pm10.62}$ & $86.67{\scriptstyle\pm5.02}$ & $33.41{\scriptstyle\pm1.41}$ \\
                                                                                        & SANE-hete   & $77.84{\scriptstyle\pm5.51}$ & $77.84{\scriptstyle\pm7.81}$  & $83.92{\scriptstyle\pm4.28}$ & $35.88{\scriptstyle\pm1.30}$ \\\cmidrule(r){2-6}
                                                                                        & \bf{Auto-HeG (ours)}  & $\bf{83.51{\scriptstyle\pm\bf{6.56}}}$ & $\bf{86.76\scriptstyle\pm4.60}$  & $\bf{87.84{\scriptstyle\pm\bf{3.59}}}$ & $\bf{37.43 {\scriptstyle\pm\bf{1.37}}}$\\\bottomrule
\end{tabular}}
%\vspace{-10pt}
\end{table*}
In summary, Auto-HeG first builds a comprehensive heterophilic search space, and then progressively shrinks the initial search space layer by layer, leading to a more compact supernet with Algo.~\ref{alg:A}. 
Finally, Auto-HeG selects the ultimate heterophilic GNN architecture with the guidance of the heterophily-aware distance in Algo.~\ref{alg:B}. 
Consequently, Auto-HeG could automatically derive powerful and expressive GNN models for learning discriminative node representations on heterophilic graphs effectively.
% to develop a heterophily-based metric as the criterion for the ultimate architecture selection of heterophilic supernet, rather than simply referring to the architecture weights.
%Therefore, we only use this strategy to compact the proposed supernet by narrowing the scope of candidate operations as the beginning step.

%However, very recent research~\cite{wang2021rethinking} has verified that using architecture weight magnitudes that reflects the operation strength selects the final model architectures might not be necessary and precise enough.
%\vspace{-0.5cm}
\section{Experiments}\label{sec:exp}
%\vspace{-0.2cm}
In this section, we first provide the experimental setting details. Then, we compare the proposed Auto-HeG with state-of-the-art human-designed and graph NAS models on the node classification task.
%, followed by the visualization and analysis of the designed architectures. 
Finally, we conduct ablation studies to evaluate the effectiveness of each component in Auto-HeG, including heterophilic search space, progressive heterophilic supernet training, and heterophily-guided architecture selection.
\begin{table*}[t]
\caption{Performance (ACC\%${\scriptstyle\pm}$std) of the proposed Auto-HeG compared with human-designed and graph NAS models on low-heterophily datasets.} %{\scriptsize{The best results are in bold and the second-best results are underline.}}}
\vspace{-7pt}
\label{tab:exp_homo}
\centering
\resizebox{0.8\textwidth}{!}{
\begin{tabular}{l|l|p{3.2cm}<{\centering}p{3.2cm}<{\centering}p{3.2cm}<{\centering}}
\toprule
Methods                                                                                 & Datasets      & Cora  & Citeseer        & Pubmed                     \\\midrule
\multirow{8}{*}{\begin{tabular}[c]{@{}l@{}}Human-designed models\end{tabular}} 
& GCN &$85.69{\scriptstyle\pm1.80}$ & $75.38{\scriptstyle\pm1.75}$     & $86.08{\scriptstyle\pm0.64}$          \\
& GAT & $86.52{\scriptstyle\pm1.41}$ & $75.51{\scriptstyle\pm1.85}$     & $84.75{\scriptstyle\pm0.51}$          \\
& GraphSAGE  &$80.60{\scriptstyle\pm3.63}$ &$67.18{\scriptstyle\pm5.46}$              & $81.18{\scriptstyle\pm1.12}$                                \\
& SGC           &$85.88{\scriptstyle\pm3.61}$ & $73.86{\scriptstyle\pm1.73}$                & $84.87{\scriptstyle\pm2.81}$                               \\
& GCNII$^{*}$ &$88.01$ & $77.13$ & $90.30$ \\
& Geom-GCN-I$^{*}$ &$85.19$ &$77.99$ &$90.05$ \\
& Geom-GCN-P$^{*}$ &$84.93$ &$75.14$ &$88.09$ \\
& Geom-GCN-S$^{*}$ &$85.27$ &$74.71$ &$84.75$ \\\midrule
\multirow{5}{*}{Graph NAS models}
& GraphNAS      & $84.10{\scriptstyle\pm0.79}$ & $68.83{\scriptstyle\pm2.09}$                & $82.28{\scriptstyle\pm0.64}$                     \\
& SNAG          & $81.01{\scriptstyle\pm1.31}$ & $70.14{\scriptstyle\pm2.40}$      & $83.24{\scriptstyle\pm0.84}$          \\
& SANE          & $84.25{\scriptstyle\pm1.82}$ & $74.33{\scriptstyle\pm1.54}$ & $87.82{\scriptstyle\pm0.57}$  \\
& SANE-hete   & $85.05{\scriptstyle\pm0.90}$ & $74.46{\scriptstyle\pm1.59}$ & $88.99{\scriptstyle\pm0.42}$  \\\cmidrule(r){2-5}
& \bf{Auto-HeG (ours)} &${86.88{\scriptstyle\pm\bf{1.10}}}$ & ${75.81{\scriptstyle\pm\bf{1.52}}}$ & ${89.29{\scriptstyle\pm\bf{0.27}}}$   \\\bottomrule
\end{tabular}
}
\vspace{-5pt}
\end{table*}
%\vspace{-0.2cm}
\subsection{Experimental Setting}
%\vspace{-0.2cm}
%\noindent\textbf{Datasets and Baselines.} 
\paragraph{Datasets and Baselines.}
We conduct experiments on seven real-world datasets with different degrees of heterophily $(\gamma_{node})$ denoted in brackets, where Cornell $(0.11)$, Texas $(0.06)$, and Wisconsin $(0.16)$ are three WebKB web-page dataset~\cite{garcia2016using}, and Actor $(0.24)$ is an actor co-occurrence network~\cite{tang2009social}. These four datasets are with high heterophily from~\cite{geom_pei2020geom}. 
Furthermore, Cora~$(0.83)$~\cite{cora_bojchevski2017deep}, Citeseer~$(0.71)$~\cite{citeseer_sen2008collective}, and Pubmed~$(0.79)$~\cite{pubmed_namata2012query} are three citation
network datasets with low heterophily.
%The detailed statistics of these datasets are as listed in Table~\ref{tab:datasets}. 
For comparisons on node classification, we take H2GCN-1 and H2GCN-2~\cite{h2gcn_zhu2020beyond}, MixHop~\cite{abu2019mixhop}, GPR-GNN~\cite{chien2020adaptive}, GCNII~\cite{chen2020simple}, Geom-GCN-I, Geom-GCN-P, and Geom-GCN-S~\cite{geom_pei2020geom}, FAGCN~\cite{fagcn_bo2021beyond}, as well as classical GCN~\cite{gcn_kipf2017semi}, GAT~\cite{velivckovic2018graph}, GraphSAGE~\cite{sage_hamilton2017inductive}, and SGC~\cite{sgc_wu2019simplifying} as human-designed model baselines for high-heterophily and low-heterophily datasets, respectively. Furthermore, we take GraphNAS~\cite{gao2020graph}, SNAG~\cite{SNAG20ZhaoWY}, and SANE~\cite{huan2021search} as graph NAS model baselines. 
To test the performance of simply modifying existing graph NAS search spaces, we develop `SANE-hete' as another baseline by directly injecting heterophilic aggregation functions into the search space of SANE~\cite{huan2021search}. 
%The classification accuracy as the metric.
%take \xin{xxx} as baselines to evaluate the effectiveness of the proposed Auto-HeG, along with the classification accuracy as the metric. \xin{details of each comparisons in appendix.}

%\noindent\textbf{Implementation.}
\paragraph{Implementation.}
All experiments run with Pytorch platform on Quadro RTX 6000 GPUs and the core code is built based on PyG library~\cite{pyg_lib_2019}. 
Following the setting in~\cite{huan2021search}, at the search stage, we search with different random seeds and select the best architecture according to the performance on validation data. 
At the train from scratch stage, we finetune hyper-parameters on validation data within fixed ranges. 
For all datasets, we use the same splits as that in Geom-GCN~\cite{geom_pei2020geom} and evaluate the performance of all models on the test sets over provided 10 splits for fair comparisons. 
We report the mean classification accuracy with standard deviations as the performance metric of node classification.
%More details can be found in \xin{Appendix?, what hyper parameters, what is their range.}. 
%For all datasets, at the search stage, we use the split of 40\%$\slash$40\%$\slash$20\% for train$\slash$valid$\slash$test; While at the train from scratch and test stage, we use the same splits with Geom-GCN~\cite{geom_pei2020geom}, \ie, 48\%$\slash$32\%$\slash$20\% for train$\slash$valid$\slash$test, to measure the performance of all models on the test sets over provided 10 splits for fair comparisons. We report the final mean classification accuracy with standard deviations.

%\vspace{-0.2cm}
\subsection{Experimental Results on Node Classification}
%\vspace{-0.2cm}
%\subsubsection{Results of Node Classification}
%The node classification results of the proposed Auto-HeG and comparison methods on high-heterophily datasets are shown in Table~\ref{tab:exp_hete}.
The node classification results of the proposed Auto-HeG and comparison methods on high-heterophily and low-heterophily datasets are shown in Table~\ref{tab:exp_hete} and Table~\ref{tab:exp_homo}, respectively. And the details of derived architectures on all datasets are provided in Appendix. 
As shown in Table~\ref{tab:exp_hete}, it can be generally observed that the proposed Auto-HeG achieves the best performance when learning on high-heterophily graphs compared with human-designed and graph NAS models. Specifically, Auto-HeG excesses the second-best model, \ie, human-designed H2GCN, with 1.35\%, 1.90\%, 1.80\%, and 1.57\% performance improvement on Cornell, Texas, Wisconsin, and Actor datasets, respectively. 
Compared with graph NAS models, first, Auto-HeG greatly outperforms all existing homophilic graph NAS models over all datasets. 
We attribute this to the elaborate incorporation of heterophily in our Auto-HeG, especially for graphs with high heterophily. 
Moreover, the superiority of Auto-HeG to SANE-hete verifies that simply injecting the heterophilic aggregation functions into the existing homophilic search space is not an effective solution.
This further illustrates the effectiveness of the proposed progressive heterophilic supernet training and heterophily-guided architecture selection in our Auto-HeG.
And incorporating the heterophily into all stages of automatic heterophilic GNN design indeed benefits learning on graphs with heterophily.

For comparisons on low-heterophily datasets in Table~\ref{tab:exp_homo},
our Auto-HeG  outperforms most human-designed classical GNN models and graph NAS models, further illustrating its ability to analyze graphs even with low heterophily. 
We attribute these results to two sub-modules in our Auto-HeG: the proposed search space which keeps many homophilic aggregation functions, and the derived progressive supernet training strategy which customizes layer-wise aggregation functions. 
These components enable the flexible construction of GNN architectures, leading to superior performance even with low-heterophily. 
%Therefore, we might conclude that even lacking the prior information on heterophily degrees, the proposed Auto-HeG could still achieve consistently impressive performance with both high-heterophily and low-heterophily, reflecting the excellent ability of our Auto-HeG to deal with complex and various heterophily.
Therefore, we could conclude that even lacking the prior information on heterophily degrees, the proposed Auto-HeG could progressively and adaptively select appropriate candidate operations along with the variation of heterophily, leading to consistently impressive performance on graphs with high-heterophily and low-heterophily.
This reflects the excellent ability of our Auto-HeG to deal with complex and various heterophily.

%\vspace{-0.2cm}
\subsection{Ablation Study}
\subsubsection{Effectiveness of heterophilic search space.}

We derive three variants of the proposed heterophilic search space to verify its effectiveness. Denoting the overall heterophilic search space as $\mathcal{O}_{{v}_{0}}$, we consider the following subsets:
%Based on the proposed heterophilic search space, we derive three variants to verify its effectiveness. Denoting the overall heterophilic search space as {\small{$\mathcal{O}_{{v}_{0}}$}}, we consider following subsets: 
(1) $\widetilde{\mathcal{O}}_{homo}$: only keep homophilic aggregation functions and remove all heterophilic ones from $\mathcal{O}_{{v}_{0}}$, to verify the importance of integrating heterophilic operations; 
(2) $\widetilde{\mathcal{O}}_{hete}$: only keep heterophilic aggregation functions and remove all homophilic ones from $\mathcal{O}_{{v}_{0}}$, to observe the performance when the search space only contains heterophilic operations;
(3) $\widetilde{\mathcal{O}}_{he\&ho}$: randomly remove several heterophilic and homophilic aggregation functions $\mathcal{O}_{{v}_{0}}$, to illustrate the effectiveness of simultaneously involving heterophilic and homophilic operations.

\begin{figure*}[t]
    \centering
    \vspace{-15pt}\includegraphics[width=0.95\textwidth]{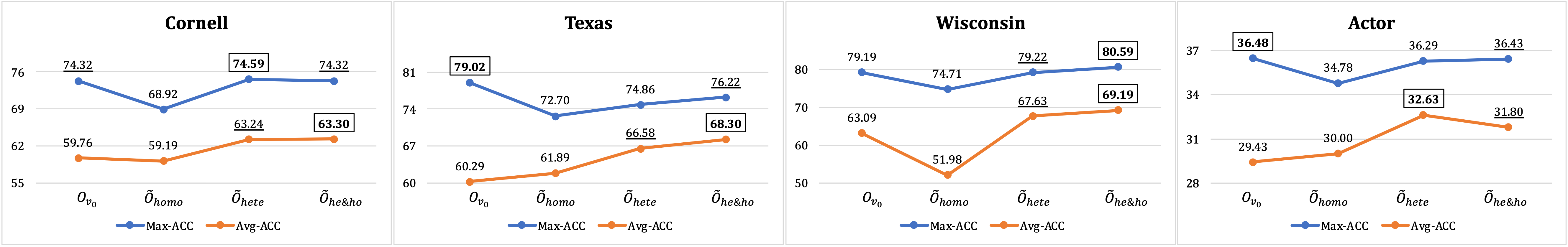}
    \vspace{-5pt}\caption{Illustration of the effectiveness of the proposed search space by evaluating search space variants.}
    \label{fig:max-avg}
\end{figure*}
\begin{figure*}[t]
    \centering
\vspace{-5pt}    \includegraphics[width=0.6\textwidth]{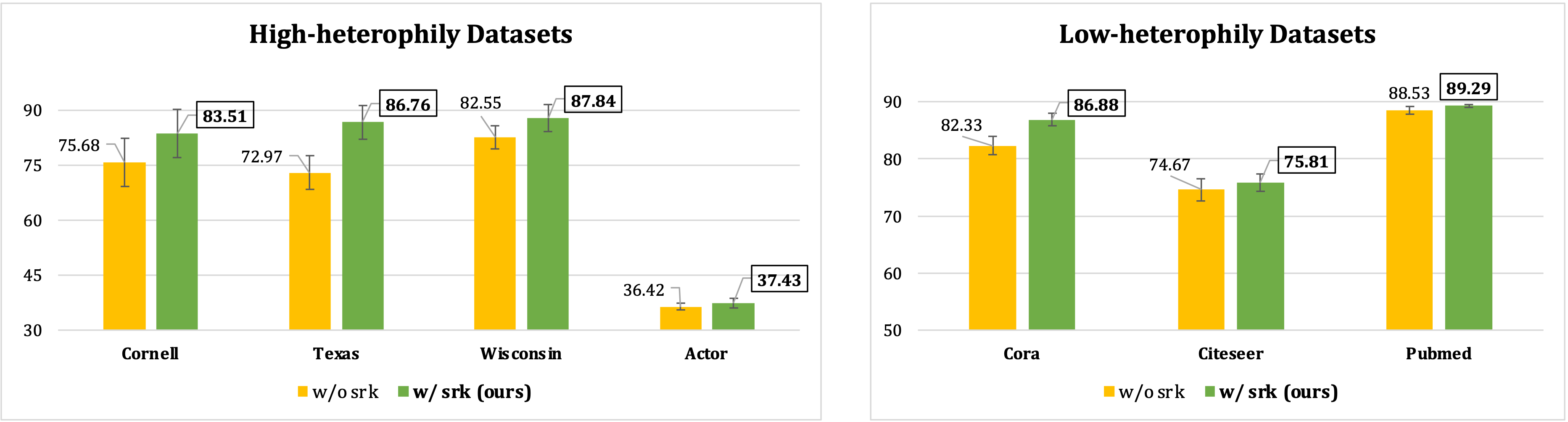}\vspace{-5pt}    
    \caption{Performance w/ and w/o shrinking (srk) for the proposed progressive supernet training.}
    \label{fig:srk}
\end{figure*}
Specifically, we randomly sample 20 architectures from each subset and report the max performance and average performance among them in Fig.~\ref{fig:max-avg}, respectively. Following observations can be obtained: 
(1) No best results are achieved in $\widetilde{\mathcal{O}}_{homo}$ on all datasets in both max and average cases, which naturally verifies the importance of heterophilic operations to heterophilic graph learning;
(2) $\widetilde{\mathcal{O}}_{hete}$ and $\widetilde{\mathcal{O}}_{he\&ho}$ generally have better performance compared to other search space sets, which illustrates the necessity of involving appropriate homophilic operations.
Moreover, this shows that merely involving heterophilic operations is not enough for complex and diverse heterophily learning;
(3) The overall search space $\mathcal{O}_{{v}_{0}}$ performs the best only on Texas and Actor datasets in the max case. This presents that $\mathcal{O}_{{v}_{0}}$ could not achieve consistently best performance under the random sampling scenario, even with the largest number of candidate operations. 
Moreover, this result also verifies that developing a more compact heterophilic search space is necessary and crucial. 
At this point, it could be concluded that the proposed progressive supernet training strategy is effective, and we could attribute this to its adaptive and dynamic layer-wise search space shrinking schema.
\begin{table*}[t]
\centering
\vspace{-5pt}
\caption{Comparison of the proposed heterophily-guided architecture selection scheme (Heter. Arch. Select.) with other architecture selection methods.}
\vspace{-5pt}
\label{tab:arch_select}
\resizebox{0.75\textwidth}{!}{
\begin{tabular}{l|cccc|ccc}
\toprule
\multirow{2}{*}{Arch. Select. Methods} & \multicolumn{4}{c|}{{\it{\footnotesize High-heterophily Datasets}}}         & \multicolumn{3}{c}{{\it{\footnotesize Low-heterophily Datasets}}} \\ \cmidrule(r){2-5}\cmidrule(r){6-8}
                                       & Cornell     & Texas       & Wisconsin   & Actor       & Cora          & Citeseer      & Pubmed       \\ \midrule
Argmax Arch. Select.                   & $79.19\scriptstyle\pm8.38$ & $79.46\scriptstyle\pm3.67$            & $86.27\scriptstyle\pm4.02$ & $37.12\scriptstyle\pm1.12$ & $85.33\scriptstyle\pm1.46$   & $75.43\scriptstyle\pm2.24$   & $88.52\scriptstyle\pm0.41$            \\
Val. Loss Arch. Select.               & $57.84\scriptstyle\pm3.67$ & $71.35\scriptstyle\pm5.30$             & $80.59\scriptstyle\pm7.09$ & $36.96\scriptstyle\pm1.08$ & $84.67\scriptstyle\pm1.64$   & $73.86\scriptstyle\pm1.11$   & $88.23\scriptstyle\pm0.53$             \\ \midrule
\bf{Heter. Arch. Select. (ours)}       & $\bf{83.51\scriptstyle\pm6.56}$ & $\bf{86.76\scriptstyle\pm4.60}$ & $\bf{87.84\scriptstyle\pm3.59}$ & $\bf{37.43\scriptstyle\pm1.37}$ & $\bf{86.88\scriptstyle\pm1.10}$   & $\bf{75.81\scriptstyle\pm1.52}$   & $\bf{89.29\scriptstyle\pm0.27}$\\ \bottomrule
\end{tabular}
}
\end{table*}
\subsubsection{Effectiveness of progressive supernet training.}
The comparison results with and without the layer-wise search space shrinking in the proposed progressive supernet training strategy are listed in Fig.~\ref{fig:srk}. 
It can be generally observed that, on both high-heterophily and low-heterophily datasets, the performance of the compact supernet consistently achieves better performance with the proposed supernet training strategy. 
For example, the compact supernet with progressive training significantly raises the performance of Cornell from 75.68\% to 83.51\% and Texas from 72.97\% to 86.76\%. 
We attribute such impressive improvements to the excellent ability of our Auto-HeG in building the effective and adaptive supernet. 
Without the progressive training, the initial search space would keep a large number of operations that might be irrelevant and redundant to specific graphs with heterophily, which brings serious challenges to the search strategy for optimizing in such a large supernet.
At this point, Auto-HeG progressively shrinks the search space layer-wisely and dynamically narrows the scope of relevant candidate operations, resulting in a more compact and effective supernet for deriving powerful heterophilic GNNs.

\subsubsection{Effectiveness of heterophily-guided architecture selection.}
We compare the proposed heterophily-guided architecture selection with the other two types of architecture selection methods, \ie, architecture weight magnitude based argmax strategy, which is the most commonly used method in existing gradient-based graph NAS~\cite{huan2021search,wang2021autogel}, and perturbation-based architecture selection method with the validation loss criterion proposed by~\cite{wang2021rethinking}. 
The comparison results are listed in Table~\ref{tab:arch_select}. 
Generally, our proposed heterophily-guided scheme with the heterophily-aware distance criterion gains the best classification performance on all datasets consistently, illustrating its effectiveness in expressive architecture selection. 
Furthermore, we can observe that the architecture weight magnitude based argmax strategy, \ie, `Argmax Arch. Select.', performs better than that of perturbation-based architecture selection method with the validation loss criterion, \ie, `Val. Loss Arch. Select.'. 
Even the work ~\cite{wang2021rethinking} has verified the effectiveness of `Val. Loss Arch. Select.' over CNN-based NAS in the leave-one-out manner, more importantly, we find that it might not well adapt to GNN-based NAS by implementing the validation loss based criterion straightforwardly. 
This fact further illustrates the effectiveness of the proposed heterophily-aware distance criterion in the leave-one-out-manner for GNN-based NAS with heterophily. 

\section{Conclusion}\label{sec:conc}
In this paper, we propose a novel automated graph neural network on heterophilic graphs via heterophily-aware graph neural architecture search, namely Auto-HeG, which is the first work of automatic heterophilic graph learning. 
By explicitly incorporating heterophily into all stages, \ie, search space design, supernet training, and architecture selection, Auto-HeG could develop powerful heterophilic GNNs to deal with the complexity and variety of heterophily effectively. 
To build a comprehensive heterophilic GNN search space, Auto-HeG includes non-local neighbor extension, ego-neighbor separation, diverse message passing, and layer-wise combination at both micro-level and macro-level. 
To develop a compact and effective heterophilic supernet based on the initial search space, Auto-HeG conducts the progressive supernet training strategy to dynamically shrink the scope of candidate operations according to layer-wise heterophily variation. 
In the end, taking the heterophily-aware distance criterion as the guidance, Auto-HeG selects excellent heterophilic GNN architectures by directly evaluating the contribution of each operation in the leave-one-out pattern. 
Extensive experiments verify the superiority of the proposed Auto-HeG on learning graphs with heterophily to both human-designed models and graph NAS models.

\section*{Acknowledgment}
This work was partially supported by an Australian Research Council (ARC) Future Fellowship (FT210100097), Guangdong Provincial Natural Science Foundation under grant NO. 2022A1515010129, and the CAS Project for Young Scientists in Basic Research (YSBR-008).
%One limitation of this work is that the proposed Auto-HeG is currently limited to the node classification task on heterophilic graphs. In future work, an interesting direction would be to extend Auto-HeG to other heterophilic analysis tasks.
\bibliographystyle{ACM-Reference-Format}
\bibliography{ref}
\clearpage
\appendix
\section{Appendix}
\renewcommand\thetable{\Alph{section}\arabic{table}} 
\renewcommand\thefigure{\Alph{section}\arabic{figure}} 
\setcounter{table}{0}
\begin{table*}[htp]
\caption{Overall comparison between the proposed Auto-HeG and existing graph NAS methods.{\scriptsize{(`Uni' and `Div' denote uniform and diverse, respectively.)}}}
\centering
\label{tab:compare}
\resizebox{0.75\textwidth}{!}{
\begin{tabular}{l|c|cccc|ccc}
\toprule
\multirow{3}{*}{Methods} & \multirow{3}{*}{Graph Types} & \multicolumn{4}{c|}{Search Spaces}                                                                                                             & \multicolumn{3}{c}{Search Strategies}                                                                                         \\ %\specialrule{0em}{1pt}{1pt}\cline{3-9}\specialrule{0em}{1pt}{1pt}
\cmidrule(r){3-9}
                         &                              & \multicolumn{3}{c|}{Micro-level}                                                                               & \multirow{2}{*}{Macro-level} & \multicolumn{1}{c|}{\multirow{2}{*}{Shrink.}} & \multicolumn{1}{c|}{\multirow{2}{*}{Optim.}} & \multirow{2}{*}{Arch. Select} \\\cmidrule(r){3-5} %\specialrule{0em}{1pt}{1pt} \cline{3-5}\specialrule{0em}{1pt}{1pt} 
                         &                              & \multicolumn{1}{c|}{Non-local Neigh.} & \multicolumn{1}{c|}{Ego-neigh. Sep.} & \multicolumn{1}{c|}{Agg.}       &                              & \multicolumn{1}{c|}{}                         & \multicolumn{1}{c|}{}                        &                               \\ \midrule
GraphNAS~\cite{gao2020graph}                 & homo.                         & \multicolumn{1}{c|}{\xmark}                & \multicolumn{1}{c|}{\xmark}               & \multicolumn{1}{c|}{Uni}        & \xmark                            & \multicolumn{1}{c|}{\xmark}                        & \multicolumn{1}{c|}{RL}                      & -                             \\ 
AGNN~\cite{zhou2019auto}                     & homo.                         & \multicolumn{1}{c|}{\xmark}                & \multicolumn{1}{c|}{\xmark}               & \multicolumn{1}{c|}{Uni}        & \xmark                            & \multicolumn{1}{c|}{\xmark}                        & \multicolumn{1}{c|}{EA+RL}                   & -                             \\ 
SANE~\cite{huan2021search}                     & homo.                         & \multicolumn{1}{c|}{\xmark}                & \multicolumn{1}{c|}{\xmark}               & \multicolumn{1}{c|}{Uni}        & \cmark                            & \multicolumn{1}{c|}{\xmark}                        & \multicolumn{1}{c|}{Dete. Differ.}           & Argmax                        \\ 
SNAG~\cite{SNAG20ZhaoWY}                     & homo.                         & \multicolumn{1}{c|}{\xmark}                & \multicolumn{1}{c|}{\xmark}               & \multicolumn{1}{c|}{Uni}        & \cmark                            & \multicolumn{1}{c|}{\xmark}                        & \multicolumn{1}{c|}{RL}                      & -                             \\\midrule
%AutoGEL~\cite{wang2021autogel}                  & homo.                         & \multicolumn{1}{c|}{\xmark}                & \multicolumn{1}{c|}{\xmark}               & \multicolumn{1}{c|}{Uni}        & \cmark                            & \multicolumn{1}{c|}{\xmark}                        & \multicolumn{1}{c|}{Stoc. Differ.}           & Argmax                        \\ \midrule
\bf{Auto-HeG (ours)}                   & hete.                         & \multicolumn{1}{c|}{\cmark}                & \multicolumn{1}{c|}{\cmark}               & \multicolumn{1}{c|}{Div} & \cmark                            & \multicolumn{1}{c|}{\cmark}                        & \multicolumn{1}{c|}{Stoc. Differ.}           & Hete-metric                   \\ \bottomrule
\end{tabular}
}
\end{table*}
In this appendix, we provide more analysis and implementation details of the proposed Auto-HeG.
Concretely, we first highlight the importance of automated heterophilic graph learning in Web. 
Then, we present a comprehensive comparison between the proposed Auto-HeG and existing graph NAS methods.
\begin{figure*}[ht]
    \centering
    \includegraphics[width=0.9\textwidth]{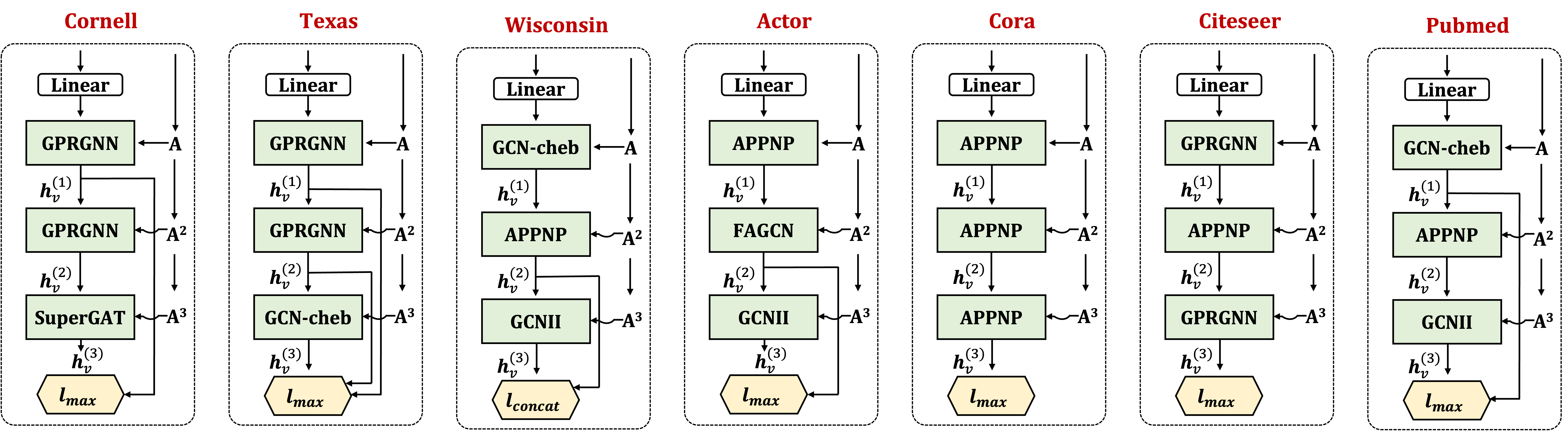}
    \caption{\small{Derived architectures by the proposed Auto-HeG.}}
    \label{fig:archs}
\end{figure*}
Next, we provide more experimental details of dataset statistics, derived architectures, search efficiency, and other complementary details. 
At last, we point out the limitation and future research directions. 
\subsection{Importance of Automated Heterophilic Graph Learning in Web}
The World Wide Web related graph learning significantly builds the connection between the massive computer understandable knowledge base and graph-structured data~\cite{craven1998learning}.
A line of research on Web-related graph neural network has greatly advanced the Web knowledge inference and information retrieve, promoting the development of Web fundamental infrastructure from the aspects of technologies and applications~\cite{geom_pei2020geom,bm_gcn_he2021block,h2gcn_zhu2020beyond,chien2020adaptive}.
However, real-world Web-based graphs are usually diverse and complex. 
And existing learning methods either develop under the homophily assumption or cost many human efforts with manual model design, incurring severe limitations to models' learning ability.
Thus, this work mainly focus on a widespread but overlooked property of Web-related graphs, {\it{heterophily}}, whose linked nodes have dissimilar features and different class labels.
Heterophily on graphs plays an important role in various real-world applications in Web ~\cite{pandit2007netprobe,zhu2021graph,h2gcn_zhu2020beyond,zheng2022graph}, such as online social networks and transaction networks, which impressively benefits the evaluating and advance of the Web socio-economic system.
We aims to model and learn such heterophily explicitly and automatically via graph neural architecture search, and tackle the critical challenges of the Web by {\textbf{developing the core Web technologies, building the automatic Web graph learning framework}}, enabling the blossoming of the {\textbf{Web as a fundamental infrastructure}} for real-world applications.
%A line of research on Web-based graph learning has greatly advanced more effective Web information retrieval and promote the applications of graph learning on Web knowledge inference~\cite{geom_pei2020geom,bm_gcn_he2021block,h2gcn_zhu2020beyond,chien2020adaptive}.
\begin{table*}[t]
\centering
\caption{Statistics of datasets.}
\label{tab:datasets}
\resizebox{0.7\textwidth}{!}{
\begin{tabular}{lp{1.5cm}<{\centering}p{1.5cm}<{\centering}p{1.5cm}<{\centering}p{1.5cm}<{\centering}p{1.5cm}<{\centering}p{1.5cm}<{\centering}p{1.5cm}<{\centering}}
\toprule
\multirow{2}{*}{Dataset} & \multicolumn{4}{c}{\it{\footnotesize High Heterophily}} & \multicolumn{3}{c}{\it{\footnotesize Low Heterophily}} \\\cmidrule(r){2-5}\cmidrule(r){6-8}
                         & Cornell   & Texas   & Wisconsin   & Actor   & Cora      & Citeseer      & Pubmed      \\\midrule
\# Nodes                 & 183     & 183     & 251     & 7600    & 2708      & 3327       & 19717      \\
\# Edges                 & 295     & 309     & 499     & 33544   & 5429      & 4732       & 44338      \\
\# Features              & 1703    & 1703    & 1703    & 931     & 1433      & 3703       & 500        \\
\# Classes               & 5       & 5       & 5       & 5       & 7         & 6          & 3          \\
\# Hete.$\gamma_{node}$   & 0.11    & 0.06    & 0.16    & 0.24    & 0.83      & 0.71       & 0.79 \\\bottomrule     
\end{tabular}
}
\end{table*}
\subsection{Technical Comparison}
We compare the proposed Auto-HeG with other existing graph NAS methods from graph types, search spaces at micro-level and macro-level, and search strategies. 

As shown in Table~\ref{tab:compare}, it can be observed that our Auto-HeG is the only method targeting heterophilic graphs. Moreover, in terms of search spaces, we consider non-local neighbor extension, ego-neighbor separation, as well as diverse aggregation functions at the micro-level, leading to better incorporation of heterophily. For search strategies, our Auto-HeG introduces the progressive shrinking scheme for supernet training and customized heterophily-aware distance metric for architecture selection, such that the supernet learning is guided by specific heterophily. In summary, all these technical differences between the proposed method and existing graph NAS guarantee our Auto-HeG outstanding ability to automated learning on graphs with heterophily.

\subsection{Auto-HeG Derived Architectures}
The detailed statistics of datasets with high-heterophily and low-heterophily are listed in Table~\ref{tab:datasets}. 
We randomly split nodes of each class into 10 random splits (48\%/32\%/20\% of nodes per class for training/validation/testing) from~\cite{geom_pei2020geom}.
And the derived architectures of all datasets are shown in Fig.~\ref{fig:archs}. 
As can be observed, different datasets desire different model architecture designs, which is reasonable and expected. 
This further demonstrates the advantage of automated graph learning on heterophilic graphs, \ie, developing customized heterophily-driven GNN architectures for different dataset.
%Besides, we notice that even low-heterophily datasets have great chances to select  
%\vspace{-8pt}
\subsection{Search Efficiency}
The search time comparison between the proposed Auto-HeG and other graph NAS models is presented in Table~\ref{tab:time}. 
We report the average time of 5 epochs for each model on all high-heterophily datasets.
Generally, the proposed Auto-HeG has comparable search time with the state-of-art graph NAS models, illustrating its excellent search efficiency. 
Specifically, Auto-HeG has $\sim3$ times search time than SANE, and such extra time cost mainly comes from the non-local neighbor extension part, since we adopt up to 3-hop neighbors in the proposed framework but SANE only used the 1-hop local neighbors. 
Note that non-local neighbor extension is the heterophily-specific design with more graph structure information. At this point, this time cost could be acceptable and would not significantly damage the search efficiency. 
Moreover, the proposed Auto-HeG still has higher search efficiency than GrapNAS model. 
In addition, although the proposed heterophily-aware architecture selection brings extra time cost, it implements only once and takes only one-round validation time. 
Hence, this time could be ignored after the all searching progress with many epochs. 
In summary, the overall comparison verifies the satisfied search efficiency of the proposed Auto-HeG.

\begin{table}[t]
\centering
\caption{Search time (clock time in seconds) comparison on high-heterophily datasets.}
\label{tab:time}
\resizebox{0.4\textwidth}{!}{
\begin{tabular}{lcccc}
\toprule
Models                  & \multicolumn{1}{l}{Cornell} & \multicolumn{1}{l}{Texas} & \multicolumn{1}{l}{Wisconsin} & \multicolumn{1}{l}{Actor} \\\midrule
GraphNAS~\cite{gao2020graph}          & 27.54                       & 55.11                     & 50.60                         & 50.63                     \\
SANE~\cite{huan2021search}              & 1.09                        & 0.79                      & 1.15                          & 0.98                      \\
\bf{Auto-HeG (ours)} & 3.26                        & 3.21                      & 3.24                          & 4.66    \\\bottomrule                 
\end{tabular}
}
\end{table}
\subsection{More Implementation Details}
%We provide more implementation details of the proposed Auto-HeG for its reproduction.

\paragraph{Search Details.} 
For the progressive supernet training, we set a three-round shrinking strategy with 200 epochs per round, and for each shrinking round, we drop the last three candidate operations, \ie, the number of candidate operations $C$ in Algo.~1 {\small{\it{(ref. main submission)}}} is $3$.
After shrinking, we further train the compact supernet for 1000 epochs for further heterophily-guided architecture selection.
For the optimization objective of architecture in Eq.~(7) {\small{\it{(ref. main submission)}}}, we set $\tau$ as the adaptive value that linearly decays with the number of epochs from maximum 8 to minimum 4. 
We use the Adam optimizer with elu activation function for architecture learning and model optimization at the search stage.

\paragraph{Train-from-scratch Details.} Since different searched architectures might have different optimal hyper-parameter spaces, we use the hyperopt tool to seek the proper hyper-parameters of the searched models for 100 iterations. Specifically, the hyper-parameter space contains: the hidden feature size with $\{16, 32, 64, 128, 256\}$, the learning rate with uniform sampling from $(1e-3, 1e-2)$, the weight decay with uniform sampling from $(1e-5,1e-3)$, the optimizer with $\{Adagrad, Adam\}$, the dropout with $\{0, 0.1, 0.2, 0.3, 0.4, 0.5, 0.6\}$, and the activation with $\{elu, relu, leakyrelu\}$. 
For fair comparison, we use the same space to fine-tune architectures when reproducing other graph NAS models' results in Table. 2 {\small{\it{(ref. main submission)}}}.

%The more detailed parameter setting of each dataset for replicating our results would be released with the code together for convenience.

\vspace{1em}
\noindent\textbf{Limitation and Future Work.}
One limitation of this work is consistent with the limitation of the current heterophilic graph learning, that is, the proposed Auto-HeG is limited to the node classification task on heterophilic graphs. In future work, an interesting direction would be to extend Auto-HeG to other heterophilic analysis tasks with more fine-grained and more elaborate designs of search spaces and search strategies.

\end{document}